\documentclass[runningheads]{llncs}

\usepackage{eccv}

\usepackage{eccvabbrv}

\usepackage{graphicx}
\usepackage{booktabs}
\usepackage[ruled,linesnumbered]{algorithm2e}
\usepackage{algorithmic}
\usepackage[accsupp]{axessibility}  

\usepackage[pagebackref,breaklinks,colorlinks,citecolor=eccvblue]{hyperref}

\usepackage{orcidlink}
\usepackage{enumitem}

\begin{document}

\title{Tuning-Free Image Customization with Image and Text Guidance} 

\titlerunning{Tuning-Free Image Customization with Image and Text Guidance}

\author{Pengzhi Li$^{1}\textsuperscript{*}$, \ \ Qiang Nie$^{2}\textsuperscript{*}$, \ \ Ying Chen$^2$,\ \ Xi Jiang$^3$,\ \ Kai Wu$^2$,\ \ Yuhuan Lin$^2$,\ \\
Yong Liu$^2$,\ \ Jinlong Peng$^2$, \ \ Chengjie Wang$^2$,\ \ Feng Zheng$^{3{\dag}}$
}

\authorrunning{Li et al.}


\institute{$^1$Tsinghua Shenzhen International Graduate School \ \
$^2$Tencent Youtu Lab \ \ \\
$^3$ Southern University of Science and Technology}

\maketitle


\begin{figure*}
  \centering
  \includegraphics[width=0.95\textwidth]{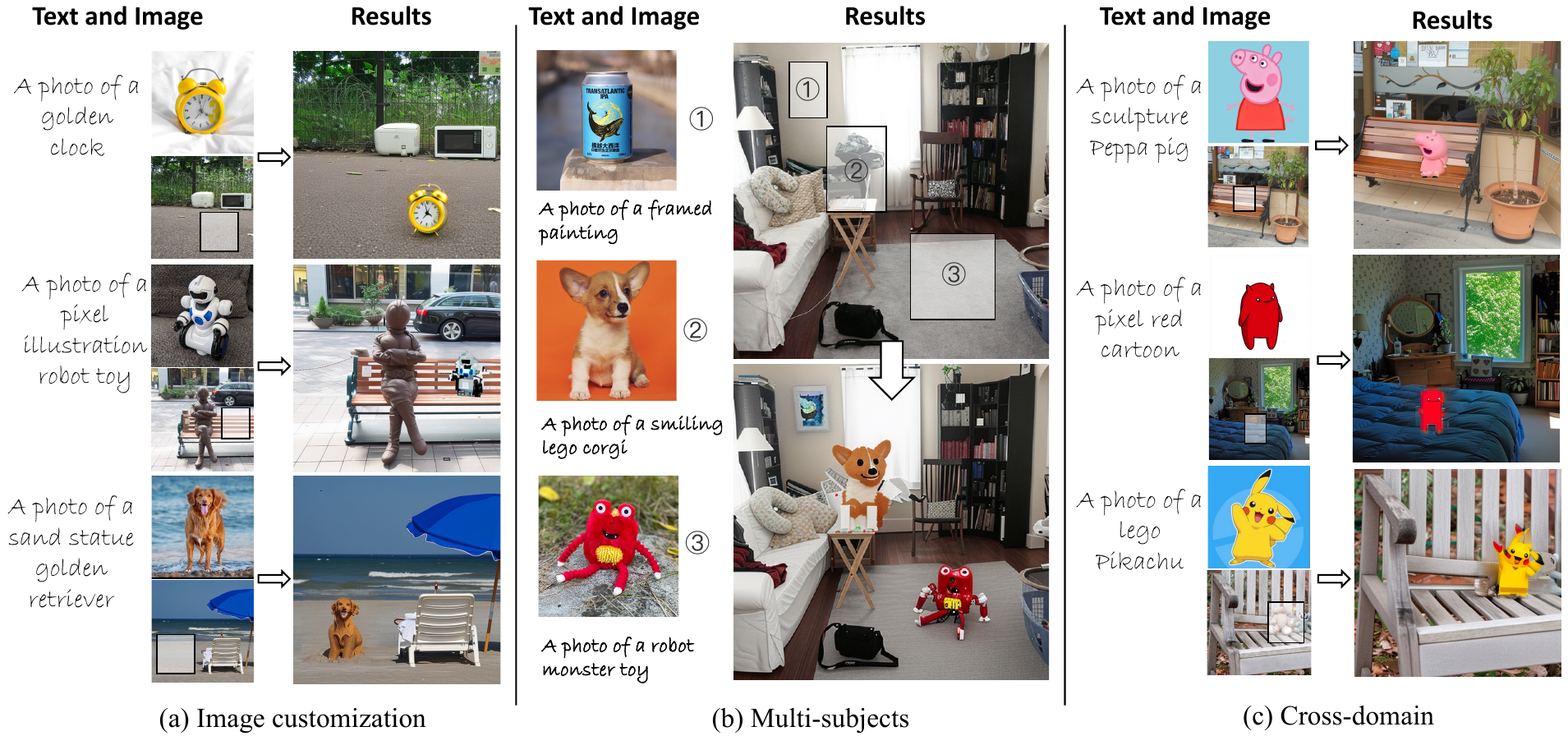}

  \caption{  Performance overview of the proposed method in image customization: (a)  The proposed method enables the generation of any subject depicted in the reference image within the designated image region to be edited. Additionally, it allows for modifying the generated subject's attributes based on the user's text description. (b) The versatility: our method can extend to scenarios involving multiple subjects from different reference images and multiple regions to be edited. (c) Cross-domain customization: driven by text, the proposed method can transform the subject in the reference image into a different domain, such as converting it into a cartoon style.
  }
 \label{fig:teaser}
\end{figure*}
\vspace{-2em}



\let\thefootnote\relax\footnotetext{
$^\dag$Corresponding author, $\textsuperscript{*}$ Equal contribution.
}

\begin{abstract}
Despite significant advancements in image customization with diffusion models, current methods still have several limitations:
1) unintended changes in non-target areas when regenerating the entire image; 
2) guidance solely by a reference image or text descriptions; and 3) time-consuming fine-tuning, which limits their practical application. 
In response, we introduce a tuning-free framework for simultaneous text-image-guided image customization, enabling precise editing of specific image regions within seconds. 
Our approach preserves the semantic features of the reference image subject while allowing modification of detailed attributes based on text descriptions. 
To achieve this, we propose an innovative attention blending strategy that blends self-attention features in the UNet decoder during the denoising process. 
To our knowledge, this is the first tuning-free method that concurrently utilizes text and image guidance for image customization in specific regions. Our approach outperforms previous methods in both human and quantitative evaluations, providing an efficient solution for various practical applications, such as image synthesis, design, and creative photography.

\end{abstract}
\vspace{-1em}

\section{Introduction}
\label{sec:intro}

Recently, with the continuous development of diffusion models~\cite{song2020denoising, dhariwal2021diffusion, rombach2022high}, there have been significant advancements in customized image generation. Given different text prompts, large-scale diffusion models such as Stable Diffusion~\cite{rombach2022high} have demonstrated the ability to generate high-quality images that align with specific text prompts. Based on Stable Diffusion, ControlNet~\cite{zhang2023adding} presents a more fancy manner to generate images according to the conditions of text description, sketch, pose, \etc. However, methods like ControlNet re-generate the entire image according to the conditions, which always causes unintended changes in non-target regions. In some cases, people may only want to edit a certain region of the image. Moreover, existing methods~\cite{mokady2023null, cao2023masactrl} heavily rely on text description for editing, which may not always capture the desired image modifications accurately even when utilizing long sentences. If given a reference image, such a kind of information misalignment can be well-tamed.
Therefore, in this paper, we are addressing the need for image customization \textit{at specified region(s)} with concurrent guidance of \textit{image} and \textit{text}.

Despite the demand, previous methods have not adequately explored the potential of using both text and images to drive image generation simultaneously. For instance,  Paint-by-example~\cite{yang2023pbe} trains a diffusion model conditioned on images, using them as templates to generate specific features in selected areas of the target image. AnyDoor~\cite{chen2023anydoor} utilizes an ID extractor to obtain ID tokens from reference images to generate subjects with consistent features. However, replacing the text embeddings with optimized image embeddings in these methods prevents pre-trained diffusion models from retaining their text-driven generative ability, which hinders the following more detailed attribute editing on target subjects. On the other hand, text-only driven approaches like BLD use text prompts to generate new subjects in selected local areas. MasaCtrl~\cite{cao2023masactrl} and Null-text Inversion~\cite{mokady2023null} use the text prompts to edit the attributions of existing foreground or background. 
While text-only driven methods retain text-editing capabilities, they are unable to generate specific new subjects based on reference images with unseen attributes.

Furthermore, many existing methods involve time-consuming dataset processing and training phases, as seen in Paint-by-example~\cite{yang2023pbe} and AnyDoor~\cite{chen2023anydoor}. Null-text Inversion~\cite{mokady2023null} requires tuning the unconditional text embedding in class-free guidance branch. Although MasaCtrl~\cite{cao2023masactrl} is implemented without tuning, its editing ability is quite limited to the subjects that have been trained in Stable Diffusion. Time efficiency is an important consideration for the scalability of the image customization method.

To address the above issues, we propose a tuning-free framework for simultaneous text-image-driven image customization, which allows users to accurately edit the specific region(s) of an image within seconds under the guidance of reference images and text descriptions, as illustrated in Fig. \ref{fig:teaser}. In the proposed framework, the content to be generated in the target region is controlled by the reference image, and the attributions of the content are edited by the text. To preserve the feature of the subject in the reference image, we create a collage by aligning the segmentation of the reference subject to the target region in the image to be edited. The collage is then inverted with DPM-Solver++ other than DDIM to obtain latent codes as the initial noise samples for the diffusion process. In the following, a three-stream denoising structure is proposed for customization. The latent code with a null prompt is utilized for image reconstruction. The latent code with text prompt is used for reference subject editing. To keep the edited subject in harmony with the image to be edited, another latent code utilized for generating the edited image is attained by filling random Gaussian noise between the reference subject area and non-target area in the latent code. The customization is achieved by a self-attention blending strategy which blends the features in reconstruction and text editing streams with target image generation stream. Since the interaction among the text, the reference image and the image to be edited is limited in the target region, our method can avoid the unintended changes in non-target area and attain precise subject attribute editing.

To our knowledge, this is the first tuning-free method that concurrently utilizes text and image guidance for specific region image customization. As shown in Fig.~\ref{fig:teaser}, our approach exhibits significant value for applications. In summary, our work makes the following contributions:

\begin{itemize}[noitemsep]
	\item We propose a tuning-free image customization framework, enabling content manipulation in the given region(s) of an image according to user-provided example images and text descriptions.
	\item We propose a self-attention blending strategy for content customization, which addresses the issue of unintended changes in non-target area in previous image editing methods and achieves precise editing of specific theme attributes.
	\item Our method outperforms previous approaches in human and quantitative evaluations, providing an efficient solution for numerous practical applications such as image synthesis, design, and creative photography.
\end{itemize}

\begin{figure*}[t]
  \centering
  \includegraphics[width=0.99\textwidth]{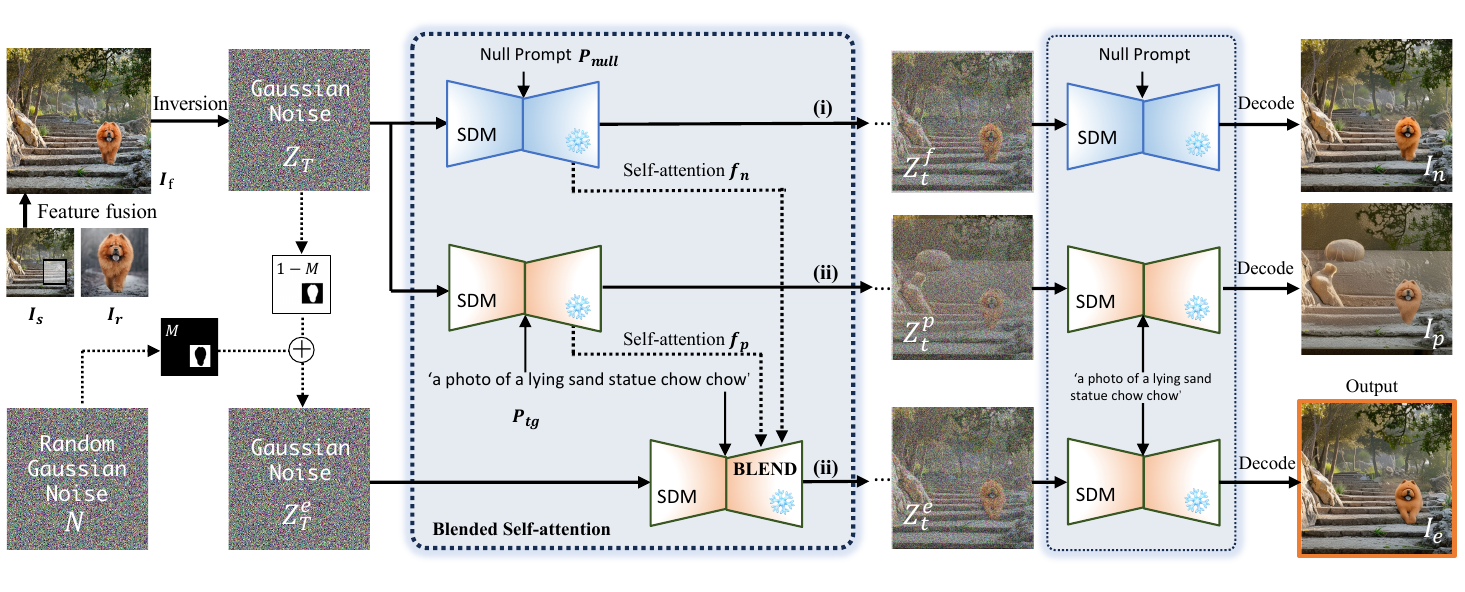} 

  \caption{ The pipeline of the proposed method. Our method uses text descriptions $T$ and the reference image $I_r$ as guidance to customize the target region(s) of the image to be edited in a tuning-free manner. We employ blended self-attention instead of original self-attention injection throughout the denoising process, which allows us to retain (i) the generated subject features while achieving (ii) the text-driven capability for attributes modification. 
  }
  \label{fig:pipeline}
\end{figure*}
\vspace{-1em}

\section{Related Work}
\paragraph{Image Editing Guided by Text/Image.} With the development of diffusion models~\cite{rombach2022high,saharia2022imagen,ramesh2022hierarchical},  image editing and generation have grown rapidly. Most previous methods~\cite{brooks2023instructpix2pix,hertz2022prompt,gal2022textual_in,chefer2023attend,couairon2022diffedit,li2023layerdiffusion,meng2021sdedit,li2023layerdiffusion} use either text descriptions or a reference image to guide image editing. Some methods~\cite{tumanyan2022plug,hertz2022prompt,meng2021sdedit,mokady2023null} focus on global editing using text descriptions, with SDEdit~\cite{meng2021sdedit} achieving editability by adding moderate noise. P2P~\cite{hertz2022prompt} and PnP~\cite{tumanyan2022plug} employ cross-attention or self-attention mechanisms for global image editing, while Null-text Inversion~\cite{mokady2023null} explores better reconstruction results during the inversion process to improve image editing. Another category of methods~\cite{avrahami2022blended,avrahami2023blended,yang2023pbe,xie2023smartbrush} concentrates on local image editing. Blended Diffusion~\cite{avrahami2022blended} and Blended Latent Diffusion~\cite{avrahami2023blended} use mask to create a blend denoising step during editing, while DiffEdit~\cite{couairon2022diffedit} can automatically generate masks during the diffusion process to achieve local editing. In contrast to the text-based approaches, Paint-by-example~\cite{yang2023pbe} trains an image-conditioned diffusion model, and AnyDoor~\cite{chen2023anydoor} uses an ID extractor for image-driven image editing. Although these works have achieved impressive results, they can only edit images based on either text or an image and learn coarse semantic information to generate low-fidelity images. Our method achieves high-fidelity image generation driven by both text and image.

\paragraph{Training for Image Customization.}
Subject-driven image editing focuses on generating content consistent with subject features in scene images. Some past customization methods~\cite{ruiz2022dreambooth,gal2022textual_in,han2023highly,kumari2023customdiff,wei2023elite,han2023svdiff,kawar2022imagic} required significant time and computational resources to fit new concept features. Dreambooth~\cite{ruiz2022dreambooth} fine-tunes diffusion models using a set of subject images, Textual Inversion employs optimized text embeddings to represent new subjects, HiPer~\cite{han2023highly} explores a set of personalized tokens to represent new subjects, and CustomDiffusion~\cite{kumari2023customdiff} captures multiple concepts simultaneously by learning new text embeddings and fine-tuning cross-attention. Break-A-Scene~\cite{avrahami2023break} can effectively learn multiple subject features. Some works~\cite{jia2023taming,chen2023anydoor,chen2023subject} have explored using the large-scale datasets for pre-training to achieve customization without fine-tuning. Although these methods can generate high-quality images, the time-consuming training phase limits their use. Training-free image generation and editing of specific subject areas remain in the exploratory stage.


\paragraph{Image Composition.}
Image composition is widely applied in various downstream tasks. A common practice in image processing is to stitch two different photos together, with many methods~\cite{xue2022dccf,cun2020improving,liu2020arshadowgan,hong2022shadow,zhang2020deep,tripathi2019learning,zhang2020learning}focusing on image harmonization to make images more realistic. These methods can generally be divided into several categories, including object placement, image blending~\cite{tripathi2019learning,zhang2020learning}, harmonization~\cite{cong2020dovenet,cun2020improving}, and shadow generation~\cite{hong2022shadow}. However, these methods struggle to change the original layout and content of the image, making the generation of images that conform to real human visual perception challenging. In this paper, we consider their feature preservation capabilities and leverage the powerful generation capabilities of diffusion models to drive the generation of realistic and harmonious images with consistent lighting and environmental features.

\section{Method}
\label{sec:formatting}
The pipeline of our method is illustrated in Fig.~\ref{fig:pipeline}. Given an image $I$ to be edited and the target region(s) $R$ that needs edition, our goal is to synthesize an image $I_e$ that not only has the subject in the reference image(s) $I_r$ but also satisfies the description of text $T$ in a tuning-free manner. The text $T$ is utilized for controlling the attributes of the customized subject in $R$.
This is a challenging task due to the following issues: 
(1) maintaining consistency in the non-target region between $I$ and $I_e$; 
(2) ensuring semantic coherence between the generated subject and the reference subject in the target region; 
(3) accurately controlling the attributes of the generated subject without changing the other part according to the text description; and 
(4) seamlessly integrating the generated subject in $R$ with the non-target region content in $I_e$.

\begin{figure*}[t]
  \centering
  \includegraphics[width=0.95\textwidth]{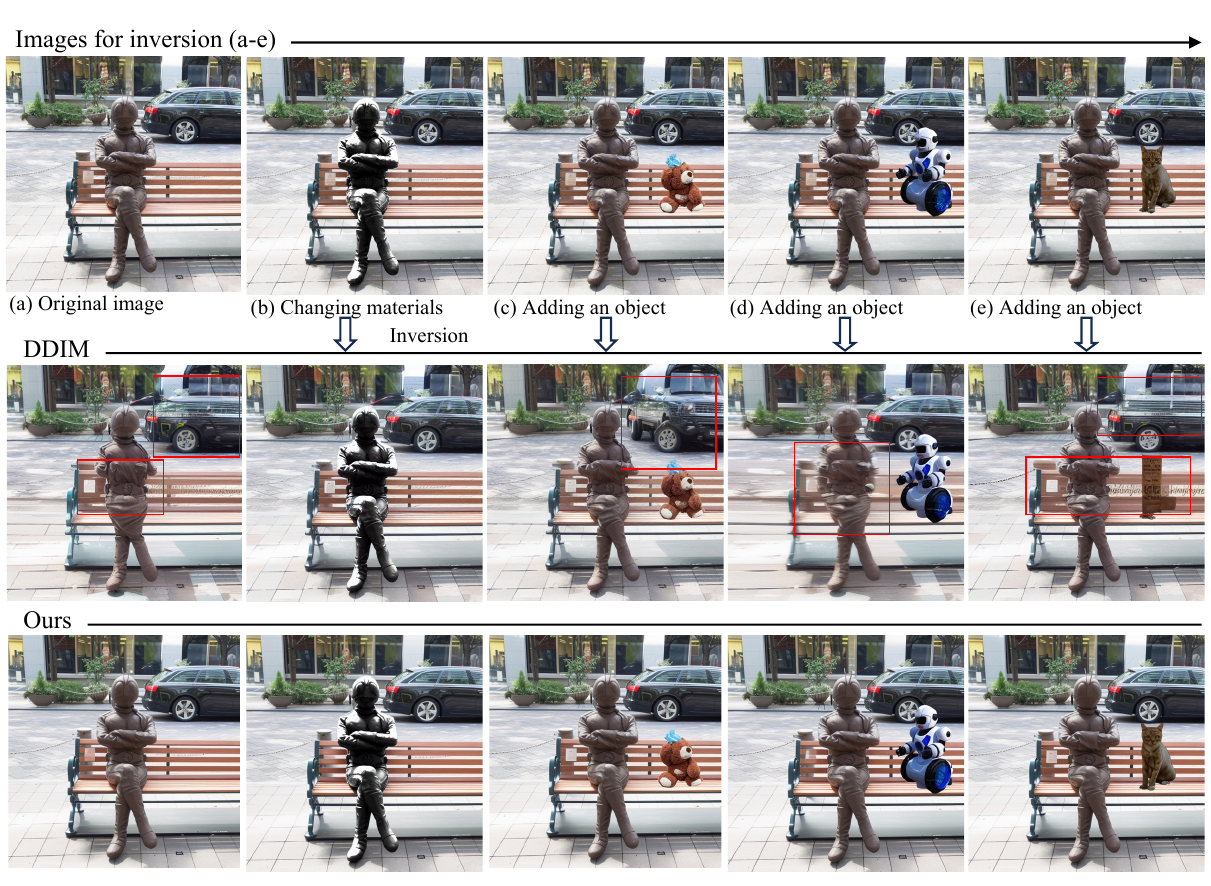}

  \caption{ Reconstruction results. The first row shows the initial images used for inversion, the second row represents the image reconstruction results from DDIM~\cite{song2020denoising}, and the third row shows our reconstruction results. DDIM's results may distort when the object's material, lighting, or additional objects in the image are artificially altered. In contrast, our method consistently generates high-quality reconstructions, a critical aspect for image editing. 
  }
  \label{fig:inversion Reconstruction}
\end{figure*}

\subsection{Image Guidance Injection and Inversion}
To retain semantic consistency, previous methods~\cite{chen2023anydoor, yang2023pbe} often encode reference image $I_r$ using a pre-trained visual encoder. However, such a kind of approach can hardly preserve the details of the reference subject $S_r$ in $I_r$. 
Different from these methods, we find that the pixels of $S_r$ contain sufficient information to keep the generated content consistent in both semantic-level and detail-level such as texture, shape and pose. 
Therefore, we directly utilize the pixels of $S_r$ into the target region $R$ for image inversion. 

To achieve aforementioned goal (1) and (2), a precise inversion process is required. The mainstream DDIM~\cite{song2020denoising} inversion effectively transforms an image into a latent representation which can successfully reconstruct the input image. However, as depicted in Fig.~\ref{fig:inversion Reconstruction}, due to differences in factors such as lighting environments between $S_r$ and $I$, the collage using conventional DDIM inversion performs less favorably compared to the real image. Inspired by~\cite{lu2023tf}, which demonstrated that utilizing high-order ODE solvers for diffusion inversion produces superior latent representations, we employ the advanced DPM-Solver++~\cite{lu2022dpm} to promote the inversion quality of the collage. As shown in Fig.~\ref{fig:inversion Reconstruction}, our approach achieves more accurate reconstruction than DDIM~\cite{song2020denoising} inversion. Therefore, in this paper, we choose to apply it to the inversion process of the diffusion model. The latent code of the collage is denoted as $z$.

Furthermore, for the goal (4), \ie the harmony between the generated subject and the non-target region in $I_e$, interaction between the reference subject in $R$ and non-target area is required. To achieve this, we fill random Gaussian noise between the reference subject and non-target area in $z$ and attain a new latent code $z^e$ to generate the final customized image. $z^e$ can be formulated as

\begin{equation}\label{ze}
    z^e = M \odot \varepsilon + (1-M) \odot z
\end{equation}

\noindent where $\varepsilon\sim N(0, I)$ is the Gaussian noise. $M$ denotes the mask to generate the region between reference subject and non-target area in the collage, as shown in Fig.~\ref{fig:pipeline}.

\begin{algorithm}[!t]    
    \caption{Proposed Method}

        \textbf{Input:} A scene image $I_s$, a reference image $I_r$, a target prompt $P_{tg}$, a null prompt $P_{null}$, the random Gaussian noise $N$, the mask $M$.\\
        \textbf{Output:} The edited image $I_e$ corresponding to $P_{tg}$.

        \begin{algorithmic}[1]
            \STATE $I_{f} = {\rm FeatureFusion}(I_s, I_r)$, $z^f_0 = {\rm Encoder}(I_{f})$ 
                \STATE ${z}_{T}^{f}\,\leftarrow...\leftarrow\,{\rm Inversion}\left(z_{0}^{f}, P_{null}\right)$
                \STATE ${{z}_{T}^{p}}\,\leftarrow...\leftarrow\,{\rm Inversion}\left(z_{0}^{f}, P_{tg}\right)$
            \STATE ${z^e_T} = {\rm FusionLatent}(z^f_T, N, M)$
            \FOR{$t = T, T-1,..., 1$}
                \STATE ${z}_{t-1}^{f}, f_{n}\,\leftarrow\,\hat\epsilon_{\theta}\left(z_{t}^{f}, t, P_{null}\right)$
                \STATE ${{z}_{t-1}^{p}}, f_{p}\,\leftarrow\,\hat\epsilon_{\theta}\left({z_{t}^{p}}, t, P_{tg}\right)$
                \STATE ${z}_{t-1}^{e}, f_{e}\,\leftarrow\,\hat\epsilon_{\theta} \left(z_{t}^{e}, t, P_{tg}\right)$
                \STATE $z_{t-1}^e \leftarrow\ {z_{t-1}^e} \cdot {M} + {z_{t-1}^f} \cdot (1 - M)$
                \STATE $f_{BD} \leftarrow {\rm BLEND}(f_{e}, f_{p}, f_{n})$
                \STATE $ z_{t-1}^{e} \leftarrow \hat\epsilon_{\theta}\left(z_{t}^{e}, P, t; f_{BD}\right),$
            \ENDFOR     

        \end{algorithmic}
        \label{alg:tic}
        \textbf{Return} $I_e = {\rm Decoder}(z_0^e)$
\end{algorithm}

\begin{figure*}[t]
  \centering
  \includegraphics[width=0.8\textwidth]{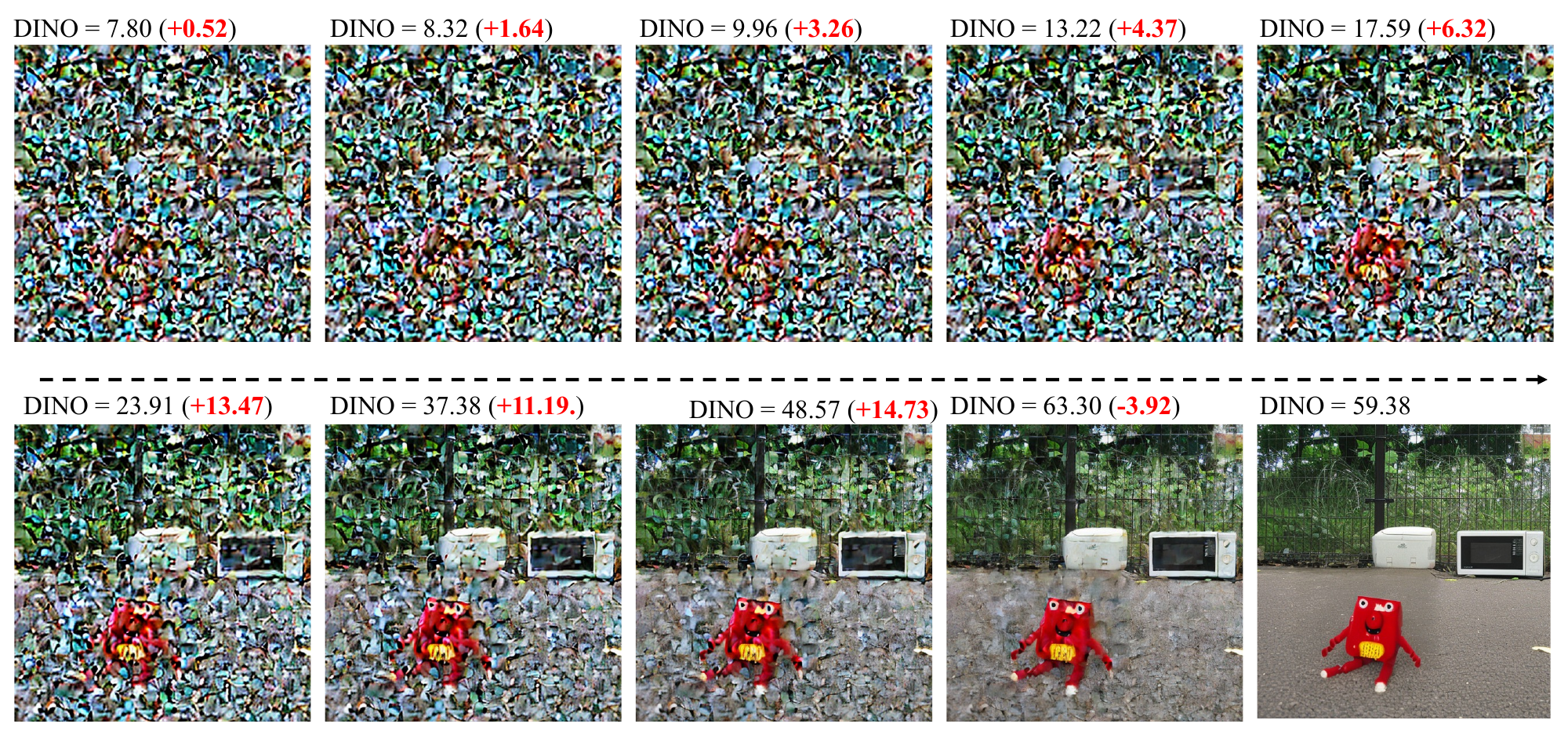} 
  \caption{ Semantic information contained in different denoising steps. We observe that the layout is mainly formed in the early denoising process, while the generation of semantic information primarily begins in the latter stages. DINO~\cite{caron2021dino} score can reflect the richness of semantic information. Therefore, we perform the attention enhancement at this stage.}
  \label{fig:semantic analysis}
\end{figure*}

\subsection{Customization with Self-Attention Blending}
To edit the attributes of generated content in target region $R$, the self-attention block in the U-Net~\cite{ronneberger2015unet} structure of the diffusion model provides a plug-and-play feature that can be seamlessly integrated into specific layers for content customization. In self-attention block, the intermediate features $f$ from the previous layer $l-1$ are projected into queries $Q$, keys $K$, and values $V$, and the output of the self-attention block can be formulated as:
\begin{equation}
\mathrm{~{Q}}=f_{t}^{l-1}{\bf W}_{l}^{q},\mathrm{\quad}{\mathrm{K}}=f_{t}^{l-1}{\bf W}_{l}^{k},\quad{V}=f_{t}^{l-1}{\bf W}_{l}^{v},
\end{equation}
\begin{equation}
    \quad \boldsymbol{A}_{t}^{l}=\operatorname{Softmax}\left(\boldsymbol{Q}_{t}^{l} \boldsymbol{K}_{t}^{l^{T}}/ {\sqrt{d}}\right),
\end{equation}
\begin{equation}
    \boldsymbol{f}_{t}^{l}=\boldsymbol{A}_{t}^{l} 
    \boldsymbol{V}_{t}^{l} \quad 
\end{equation}

\noindent where $\boldsymbol{A}$ is the attention map. The attention map contains rich structural and content information. Manipulation in self-attention layers requires no additional optimization, allowing users to achieve image recreation within seconds effortlessly.

Specifically, as illustrated in Fig.~\ref{fig:pipeline}, we utilize a three-stream architecture to execute the self-attention blending. Given an input image, we first obtain the latent representation $z^{n}$ of the collage after feature inversion. Subsequently, at each time step $t$, we pass the latent code $z$ to a denoising U-Net using the null and target text descriptions, respectively. The output feature of the self-attention block in these two streams are $f_n$ and $f_p$. Similarly, $f_e$ can be achieved from $z^e$ stream. $f_n$ from the reconstruction stream helps retain the information of non-target region. $f_p$ from the text-driven stream provides the information for attribute editing. $f_e$ offers diversity with additional Gaussian noise for the interaction between the generated subject and the non-target area. We then blend the self-attention, denoted as $f_{BD}$, using the most straightforward weighted average, as show in Eq.~\ref{eq:blended}. 

\begin{equation}\label{eq:blended}
    f_{BD} = \left\{ \begin{array}{ll}
           \alpha f_e + \beta f_p + \gamma f_n & \text{if } t \in (\tau_a T, \tau_b T)\\
           \frac{1}{2}(f_e + f_p) & \text{if } t > \tau_b T \\
           f_e & otherwise
         \end{array}\right.
\end{equation}
\noindent where $\alpha + \beta + \gamma=1$. This core operation maintains semantic information consistency while enabling text guided capabilities. Finally, we inject $f_{BD}$ during the specific denoising steps of $z_{t}^{e}$:
\begin{equation}
     z_{t-1}^{e}=\hat\epsilon_{\theta}\left(z_{t}^{e}, P, t; f_{BD}\right),
\end{equation}
where $\hat\epsilon_{\theta}$ represents the modified denoising step with $f_{BD}$. Therefore, in time step $t-1$, the self-attention block is calculated using $f_{BD}$.

\begin{equation}
\scalebox{0.85}{$
\mathrm{{Q}_{t-1}}=f_{BD}{\bf W}^{q},\mathrm{\quad}{\mathrm{K}_{t-1}}=f_{BD}{\bf W}^{k},\quad{V}_{t-1}=f_{BD}{\bf W}^{v},
$}
\end{equation}

\begin{figure*}[t]
  \centering
  \includegraphics[width=0.99\textwidth]{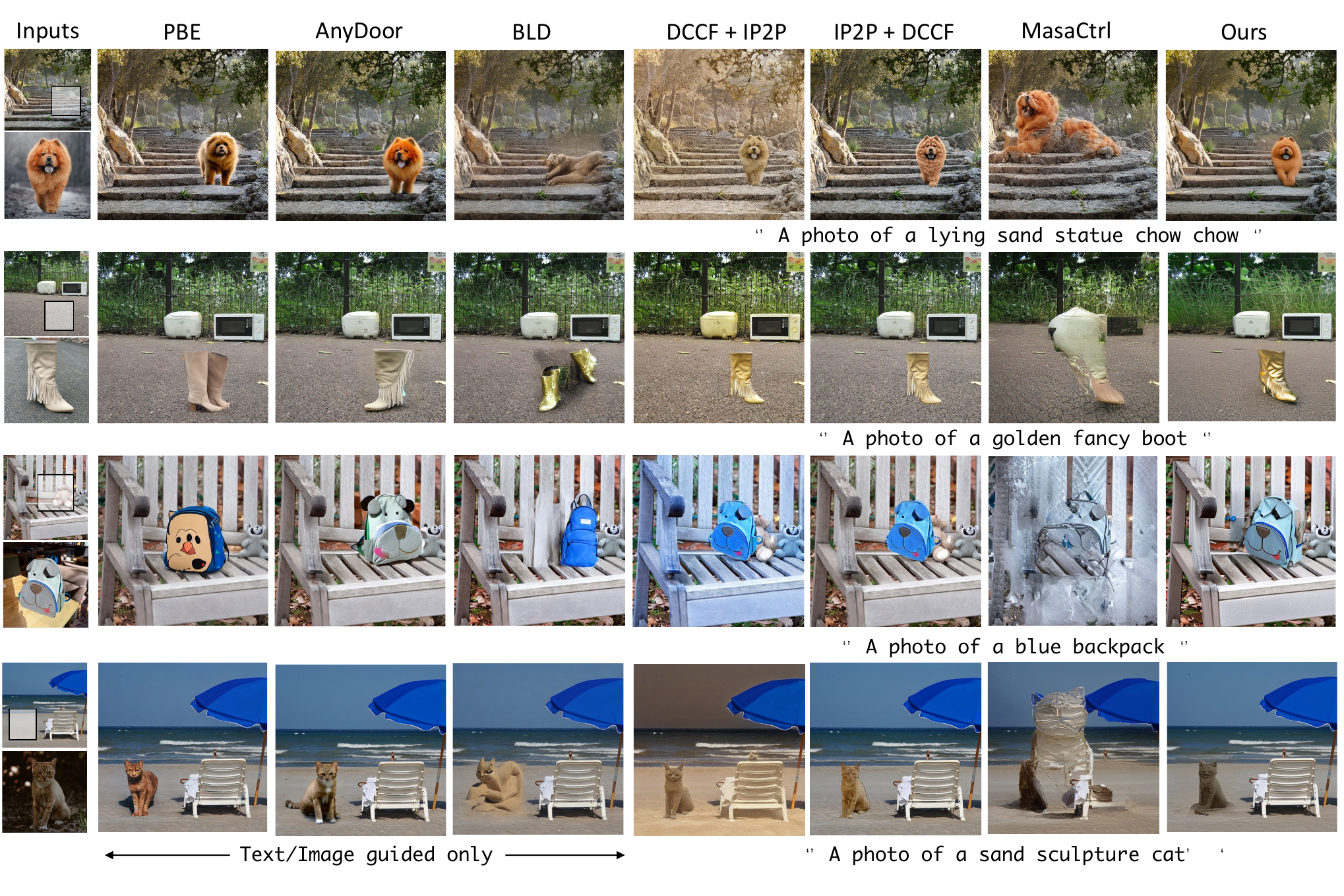} 
  \vspace{-1em}
  \caption{ Qualitative comparison with existing state-of-the-art methods. PBE~\cite{yang2023pbe} and AnyDoor~\cite{chen2023anydoor} are methods guided only by images, while BLD~\cite{avrahami2023blended} uses text as the only guidance. To evaluate the efficiency of our method, we set up an additional group of two-step methods, including first using image stitching and harmonization followed by text guided image editing (DCCF~\cite{xue2022dccf} + IP2P~\cite{brooks2023instructpix2pix}, MasaCtrl~\cite{cao2023masactrl}) and another method involving editing first and then harmonizing (IP2P~\cite{brooks2023instructpix2pix} + DCCF~\cite{xue2022dccf}). These methods can only focus on text or image, global or local editing. Our method outperforms all these methods and overcomes their limitations, achieving text and image guided local editing and generation.}
  \label{fig:Quantitative comparisons}
  \vspace{-1em}
\end{figure*}

\paragraph{Blended Enhancement}
While the proposed blending self-attention manipulation simply but effectively integrates target semantic information into the image structure, it can still result in inaccurate edition in output image and the unintended content. We believe this issue stems from the overfitting of global semantic information. Given that there's ample semantic information in the late diffusion denoising stages, further injection of blended self-attention may lead to more artifacts. To address this, we establish a threshold to determine when to cease injecting blended self-attention during the denoising steps.

We draw inspiration from~\cite{zhang2023prospect} that the diffusion model denoising U-Net generates images in the order of "layout → content → material/style. Specifically, only at a specific time step ($\tau_a$, $\tau_b$), we pass the latent representations $f_{t}^{p}$ and $f_{t}^{n}$ to the denoising U-Net of $z_T^e$, as shown in the first row of eq.~\ref{eq:blended}. When time step $t$ larger than $\tau_b T$, we only blend the $f_{t}^{p}$ with $f_{t}^{e}$ to inject more information from the text-guided stream for better attributes modification in the target region, as shown in second row of eq.~\ref{eq:blended}. For early stage before $\tau_b T$, we don't apply any blending to avoid affecting the layout generation. 
This enhancement strategy effectively corrects the inaccuracy of semantic information in target region, improving the overall quality of text-image-guided editing in the final output. 

\begin{algorithm}[!t]    
    \caption{BLEND}
        \textbf{Input:} $f_e$, $f_p$, $f_n$. \\
        \begin{algorithmic}[1]
        \STATE \textbf{if}~{$t > \tau_aT$ and $t <\tau_bT$}\\ ~~~~~\textbf{then}{~$f_{BD} \gets \alpha f_e + \beta f_p + \gamma f_n$}\\
        \textbf{else if}~{$t > \tau_bT$}~~\textbf{then}{~$f_{BD} \gets \frac{1}{2}(f_e + f_p)$}
        ~~\textbf{else}~{~$f_{BD} \gets f_e$}

        \end{algorithmic}
        \label{alg:Blend}
        \textbf{Return} $f_{BD}$
\end{algorithm}

To determine the appropriate threshold range for parameters $\tau_a$ and $\tau_b$, we conduct an analysis of the generated images and observe that altering semantic information at early or late stages can deteriorate the final results. 
As shown in Fig.~\ref{fig:semantic analysis}, for our denoising process, semantic information (quantitatively represented by DINO score) appears infrequently between ($0$, $0.5 \times T$), and rapidly increases starting at ($0.5 \times T$, $0.8 \times T$). In the next section, we conduct detailed experimental analysis on the two thresholds. The whole framework of our algorithm is illustrated in Alg. 1. The BLEND function is illustrated in  Alg. 2

\section{Experiments}
\subsection{Experimental Setup}
\paragraph{Benchmarks}

Since there is no existing dataset to evaluate specified region customization with both text and image inputs, we collect a dataset comprising 3000 images for quantitative evaluation. The sample images in this dataset span 30 categories from DreamBooth~\cite{ruiz2023dreambooth}, including 21 objects and nine living subjects. We select ten representative scenes from the COCO dataset~\cite{lin2014coco}, covering indoor and outdoor environments, and provided corresponding bounding box information. Subsequently, we generate ten sets of text for attribute modification and applied them to scenes across all categories. We provide more details in the supplementary material.

\paragraph{Evaluation metrics}

Our approach employs a dual-driven mechanism using both images and text, necessitating evaluation from both textual and visual perspectives. Our metrics is same as Dreambooth~\cite{ruiz2023dreambooth}, prioritize subject fidelity. This involves ensuring that the generated images maintain consistency with the reference subject's features. To achieve this, we utilize CLIP-I and DINO~\cite{caron2021dino} metrics to calculate the similarity of subject features within the edited regions. The second metric assesses the consistency between the edited regions and textual descriptions. Furthermore, we employ the CLIP-T metric to measure the cosine similarity between text prompts and CLIP~\cite{radford2021clip} embeddings. Additionally, we conduct user studies to comprehensively evaluate the feasibility of our approach.

\begin{table}[t]
\centering
\caption{Quantitative comparison of different methods. We report three scores: DINO~\cite{caron2021dino} score, CLIP-I, and CLIP-T, which are used to comprehensively evaluate the similarity of subject features and the matching degree of text descriptions. Our method achieves the best scores on all three metrics.
\label{table:comparison_single}}
\resizebox{.65\linewidth}{!}{
  \begin{tabular}{lcccc}
    \toprule
    Method & DINO $\uparrow$ & CLIP-I $\uparrow$ & CLIP-T $\uparrow$\\
    \midrule
    Paint-by-example~\cite{yang2023pbe} & 15.26 & 66.94 & 20.62 \\
    Stable Diffusion Inpainting~\cite{rombach2022high} & 14.43 & 62.17 & 21.04 \\
    Blended Latent Diffusion~\cite{avrahami2023blended} & 17.17 & 63.21 & 21.14 \\
    Ours & \textbf{51.18} & \textbf{78.08} & \textbf{26.86} \\
    \bottomrule
  \end{tabular}
}
\end{table}

\begin{table}[h]
\centering
\caption{Quantitative comparison of different methods. Left: Text-driven capabilities. Right: single-driven comparisons. 
\label{table:anydoor}}
\resizebox{0.85\columnwidth}{!}{
  \begin{tabular}{lcc}
    \toprule
    Method  & CLIP-T $\uparrow$\\
    \midrule
    Paint-by-example~\cite{yang2023pbe} & 20.62 \\
    Anydoor ~\cite{chen2023anydoor} & 21.22 \\
    Ours & \textbf{26.86} \\
    \bottomrule
  \end{tabular}

    \begin{tabular}{lcccc}
    \toprule
    Method & DINO $\uparrow$ & CLIP-I $\uparrow$ \\
    \midrule
    Paint-by-example~\cite{yang2023pbe} & 15.26 & 66.94  \\
    Anydoor ~\cite{chen2023anydoor} & 59.19 & 78.92  \\
    Ours & \textbf{62.88} & \textbf{80.28} \\
    \bottomrule
  \end{tabular}
}
\vspace{-0.6em}
\end{table}

\begin{table}[t]
\centering
\caption{Quantitative comparison of reconstruction results. We reconstruct a higher quality image compared to DDIM~\cite{song2020denoising}.
\label{table:inversion}}
\resizebox{.35\linewidth}{!}{
  \begin{tabular}{lcccc}
    \toprule
    Method & MAE $\downarrow$ & LPIPS $\downarrow$ & SSIM $\uparrow$\\
    \midrule
    DDIM & 0.128& 0.472 & 0.697 \\
    Ours & \textbf{0.041} & \textbf{0.106} & \textbf{0.806} \\
    \bottomrule
  \end{tabular}
}

\end{table}


\subsection{Comparison with Previous Works}
\paragraph{Single-driven method}

As shown in Fig.~\ref{fig:Quantitative comparisons}, we present the generated results of the currently outstanding single-driven methods. The Paint-by-example~\cite{yang2023pbe} utilizes example images as references to generate images in the selected regions of the scene image, matching the features of the example image. However, it can only learn coarse semantic information and fails to capture the features that best reflect the details of the subject. Based on text descriptions, Blended Latent Diffusion~\cite{avrahami2023blended} can generate corresponding images in the target area. We input as detailed text descriptions as possible, but due to the limited training samples of diffusion models, they can only generate rough object categories. In contrast, our method produces subject features highly consistent with both the example image and text descriptions. We provide a quantitative evaluation of these methods in Tab.~\ref{table:comparison_single}. We also include quantitative comparisons with Anydoor in Tab.~\ref{table:anydoor}. It's important to highlight that our method excels in retaining text-driven capabilities, while Anydoor falls short in this aspect, relying solely on reference images. Furthermore, Anydoor requires the collection of 14 datasets and involves significant training time compared to our effective tuning-free method.
\paragraph{Two-steps method}
Due to the absence of a one-step, dual-driven generation method capable of simultaneously incorporating both text and images, we adopt a two-step generation strategy for comparison. The first strategy involves segmenting the subject and seamlessly integrating it into the scene image. This is achieved using the harmonization algorithm DCCF~\cite{xue2022dccf} and the text-driven image editing algorithm IP2P~\cite{brooks2023instructpix2pix}, resulting in the outcomes (DCCF~\cite{xue2022dccf} + IP2P~\cite{brooks2023instructpix2pix}, DCCF~\cite{xue2022dccf} + MasaCtrl~\cite{cao2023masactrl}). The second strategy entails editing the example image first and then proceeding with integration and harmonization (IP2P~\cite{brooks2023instructpix2pix} + DCCF~\cite{xue2022dccf}). These methods require intricate procedures and face challenges in achieving harmonious interactions between the background and subject. In contrast, our approach accomplishes the task in a single operation and generates higher-quality images.

\begin{table}[t]
\centering
\caption{Quantitative ablation studies on the core components of our method. The values of $\alpha$ and $\beta$ represent the different order of attention blending processes, resulting in different weights.
\label{table:ablation_experiment}}
\resizebox{.8\linewidth}{!}{
  \begin{tabular}{lcccc}
    \toprule
    Method & DINO $\uparrow$ & CLIP-I $\uparrow$ & CLIP-T $\uparrow$\\
    \midrule
    Baseline & 32.30 & 73.88 & 25.93 \\
    + Blended Self-attention & 50.43 & 77.73 & 26.16 \\
    + Enhancement($\alpha$ = 1/4, $\beta$ = 1/4) & 50.71 & 77.93 & 26.44 \\
    + Enhancement($\alpha$ = 1/3 , $\beta$ = 1/3) & 47.90 & 77.75 & 26.35 \\
    + Enhancement($\alpha$ = 1/4 , $\beta$ = 1/2) & \textbf{51.18} & \textbf{78.08} & \textbf{26.86} \\
    \hline
    Blended $\rightarrow$ $(0, T)$ & 47.32 & 75.96 & 24.97 \\
    Enhancement $\rightarrow$ $(0, T)$ & 39.08 & 74.37 & 24.85 \\
    Enhancement $\rightarrow$ $(0,T/2)$ & 50.34 & 77.80 & 26.31 \\

    \bottomrule
  \end{tabular}
}


\end{table}

\begin{table}[t]
\centering
\caption{Result of user study. Our method achieves the highest scores.}
\vspace{-0.5em}

\resizebox{0.7\columnwidth}{!}{
  \begin{tabular}{lcccc}
    \toprule
    Method & Fidelity $\uparrow$ & Quality $\uparrow$ & Text alignment $\uparrow$\\
    \midrule
    Paint-by-example~\cite{yang2023pbe} & 3.46 & 3.49 & 2.88 \\
    Blended Latent Diffusion~\cite{avrahami2023blended} & 2.93 & 3.55 & 3.88 \\
    \hline
    DCCF~\cite{xue2022dccf} + IP2P~\cite{brooks2023instructpix2pix}  & 3.63 & 3.55 & 3.88 \\
    IP2P~\cite{brooks2023instructpix2pix} + DCCF~\cite{xue2022dccf}  & 3.91 & 3.31 & 4.11 \\
    MasaCtrl~\cite{cao2023masactrl}  & 2.33 & 2.34 & 2.52 \\
    Ours & \textbf{4.02} & \textbf{3.93} & \textbf{4.28} \\
    \bottomrule
  \end{tabular}
}

\label{table:user study}
\end{table}

\subsubsection{User Study}

We conduct an user study to compare our work in detail with previous methods. We select six methods for evaluation, including Paint-by-example~\cite{yang2023pbe}, Belended Latent Diffusion~\cite{avrahami2023blended}, DCCF~\cite{xue2022dccf} + IP2P~\cite{brooks2023instructpix2pix}, IP2P~\cite{brooks2023instructpix2pix} + DCCF~\cite{xue2022dccf}, and DCCF~\cite{xue2022dccf} + Masactrl~\cite{cao2023masactrl}. For each method, we generate 20 groups from 120 different images. Each set of images included additional scene images, text descriptions, and example images. Clear rules are established for evaluating fidelity, quality, and text alignment, with scores ranging from 1 to 5, representing the worst to the best. Fidelity is designed to evaluate the similarity of image features, quality to judge the harmony of images, and text alignment to evaluate whether the generated subjects within the region matched the text descriptions. We collected a total of 2513 valid answers. As shown in Tab.~\ref{table:user study}, our method achieves outstanding scores on all metrics.

\begin{figure*}[t]
  \centering
  \includegraphics[width=0.99\textwidth]{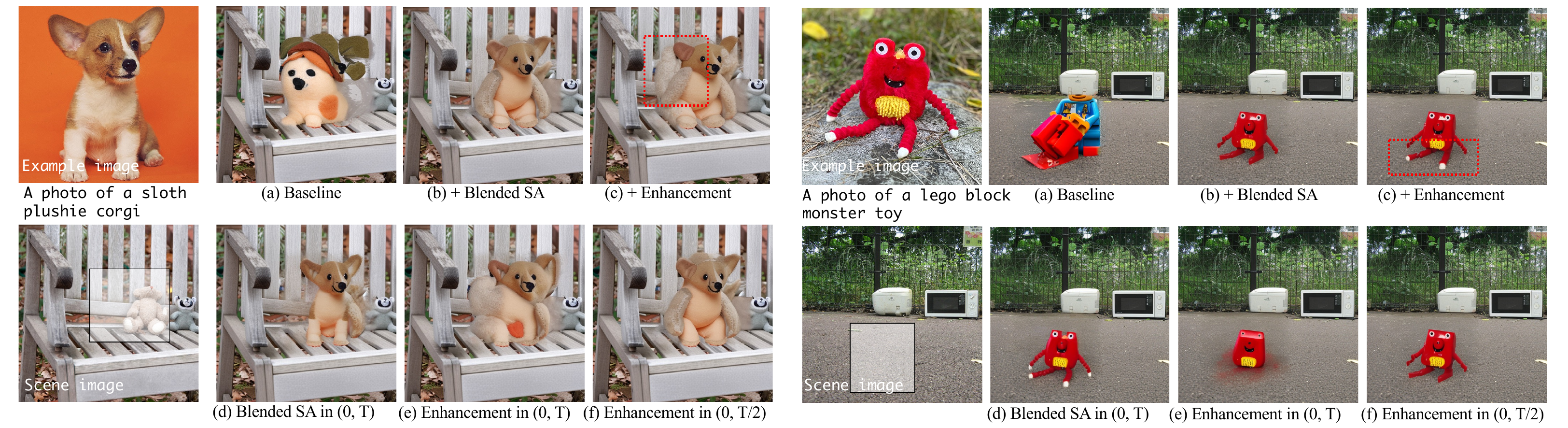} 

  \caption{Ablation studies of each components. (a) is the baseline with only inversion performed. (b) represents the blended self-attention method, while (c) adds the enhancement strategy. (d), (e) and (f) show the results with different threshold values. }
  \label{fig:ablation}
\end{figure*}
\subsection{Ablation Studies}

We conduct an extensive ablation study to validate the effectiveness of our designed core components. 
\paragraph{Inversion.} Firstly, we verify the effectiveness of the DPM-Solver++~\cite{lu2022dpm} solver in inversion inharmonious lighting images after feature fusion. In comparison to the DDIM~\cite{song2020denoising} method, our approach demonstrates superior performance. We employ 300 sets of scene images with different subjects for validation, as illustrated in Fig.~\ref{fig:inversion Reconstruction} and Tab.~\ref{table:inversion}, DDIM~\cite{song2020denoising} exhibits varying degrees of image distortion due to significant differences between the pasted subject and the original image. Experimental results show that our method outperforms DDIM~\cite{song2020denoising} significantly in metrics such as MAE, LPIPS~\cite{zhang2018unreasonable}, and SSIM, designed to assess image generation quality.
\paragraph{Blended self-attention.} As shown in Tab.~\ref{table:ablation_experiment}, we conduct comprehensive ablation experiments on the blended self-attention method to validate its effectiveness. (a) Baseline: No attention injected, only inversion performed. (b) Integration of blended attention. (c) Integration of blended enhancement strategy. (d) Setting the threshold of blended attention to 1. (e) Setting the threshold of enhancement strategy to (0, 1). (f) Setting the threshold of the enhancement strategy to (0, 0.5). In Fig.~\ref{fig:ablation}, we also provide visual results of generated images under different settings. These results strongly support the effectiveness of our blended self-attention method. By applying blended self-attention and enhancement strategies, our approach achieves a significant improvement.


\begin{figure*}[t]
  \centering
  \includegraphics[width=0.95\textwidth]{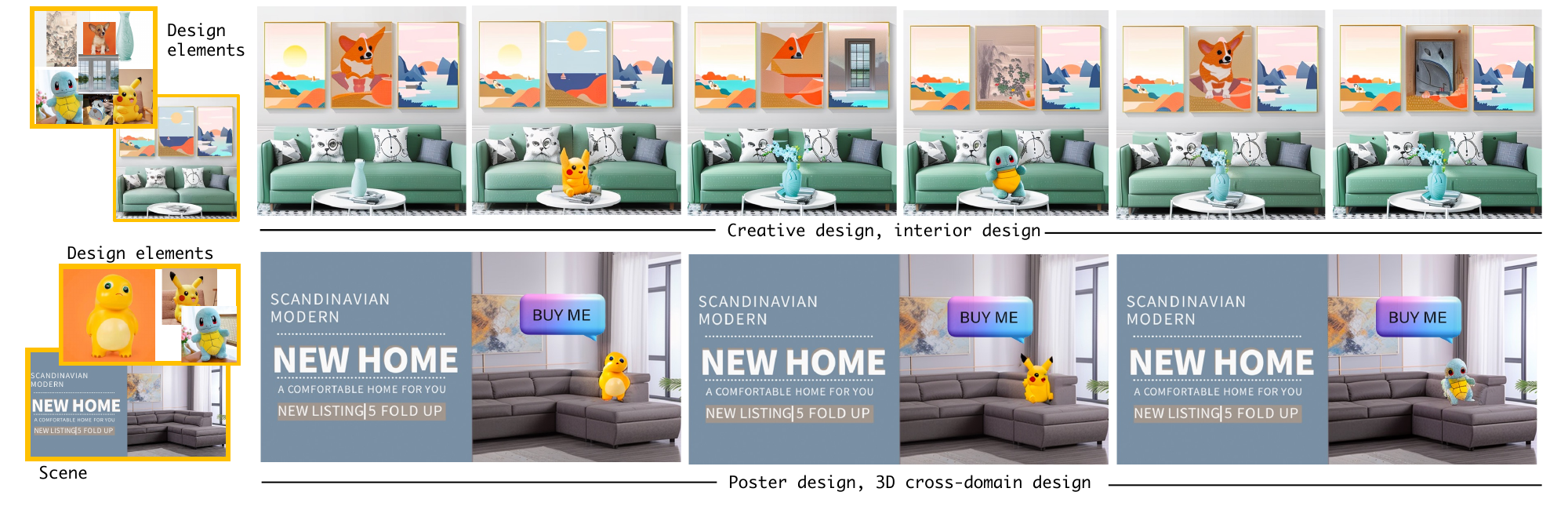} 


    \caption{Some creative applications. As shown in the first row, given an indoor scene and a collection of materials, our method can edit the interior decorations and furnishings using reference subjects from the material library. Our method can also be applied to cross-domain graphic design creations, as shown in the second column, where cartoon characters are generated directly in real-world scenes.}

  \label{fig:application}
\end{figure*}

\begin{figure*}[ht]
  \centering
  \includegraphics[width=0.85\textwidth]{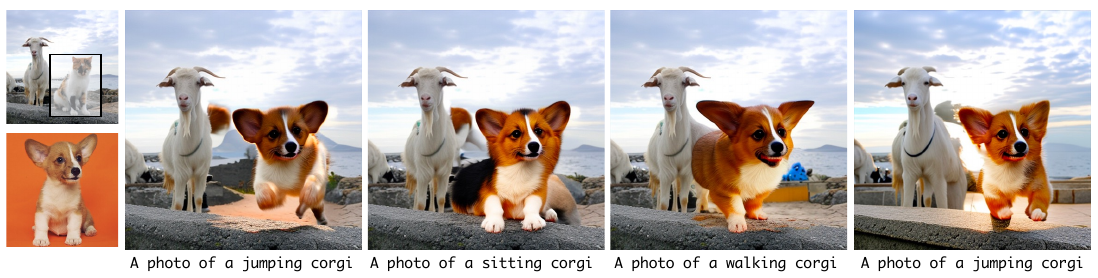} 
  \caption{Non-rigid and perspective editing may sometimes loss the subject features.}
  \label{fig:failure}
\vspace{-1mm}
\end{figure*}

\section{Application}
\paragraph{Creative photography}

Post-production in photography is crucial, and inserting or editing objects in photos has excellent potential. Relying on traditional image processing software like Photoshop requires significant time to adjust the attributes of inserted objects to make them harmonious with the environment. Fig.~\ref{fig:teaser} (a) and (b) show that our method can achieve text and example image dual-driven post-production within seconds.
\vspace{-4mm}
\paragraph{Graphic design}

As shown in Fig.~\ref{fig:application}, we demonstrate that our method can provide strong support for many creative design applications, such as interior design, poster design, and more. 
\vspace{-1em}
\section{Limitations}

Our method employs a simple yet effective hybrid strategy that maintains the subject's characteristics while possessing text-driven capabilities. However, due to self-attention blending mechanism without tuning, as other tuning-free methods, generating images from multiple perspectives is still challenging. As shown in Fig.~\ref{fig:failure}, editing non-rigid motion can also result in losing subject features. This issue has long troubled the field and urgently needs to be addressed.
\vspace{-1mm}
\section{Conclusion}
We introduce a novel tuning-free framework for image customization that effectively leverages both text prompts and reference images. Our innovative blended self-attention strategy ensures precise editing while enabling us to maintain generated subject features and simultaneously achieve text-driven capability. As a pioneering approach in this domain, it demonstrates superior performance in evaluations and provides an efficient, versatile solution for a wide range of practical applications.

\bibliographystyle{splncs04}
\bibliography{main}
\end{document}